\RequirePackage{snapshot}
\documentclass[10pt,twocolumn,letterpaper]{article}

\usepackage{cvpr}
\usepackage{times}
\usepackage{epsfig}
\usepackage{graphicx}
\usepackage{amsmath}
\usepackage{amssymb}

\usepackage{caption}
\DeclareCaptionFont{ninept}{\fontsize{9pt}{11pt}\selectfont #1}
\usepackage{adjustbox}
\usepackage{comment}
\usepackage{epigraph}
\usepackage{subcaption}
\usepackage{booktabs}

\usepackage[usenames, dvipsnames]{color}
\usepackage{array}
\usepackage{footnote}
\makesavenoteenv{tabular}
\makesavenoteenv{table}

\newcommand{\etalWspace}[0]{{\it et al.\ }}
\newcommand{\methodTag}[0]{ActionVLAD}
\newcommand{\tableSize}[0]{\scriptsize}
\definecolor{ao(english)}{rgb}{0.0, 0.5, 0.0}
\newcommand{\old}[1]{{\color{red} #1}}
\newcommand{\new}[1]{{\color{ao(english)} #1}}
\newcommand{\oldnew}[2]{\old{#1} $\rightarrow$ \new{#2}}

\usepackage[hyphens]{url}
\usepackage[pagebackref=true,breaklinks=true,letterpaper=true,colorlinks,bookmarks=false]{hyperref}

\cvprfinalcopy 
\def\cvprPaperID{327} 

\ifcvprfinal\fi
\begin{document}

\title{\methodTag{}: Learning spatio-temporal aggregation for action classification}

\author{Rohit Girdhar$^{1\thanks{Work done at Adobe Research during RG's summer internship}}$
\quad
Deva Ramanan$^{1}$
\quad
Abhinav Gupta$^{1}$
\quad
Josef Sivic$^{2,3\footnotemark[1]}$
\quad
Bryan Russell$^{2}$
\\
$^{1}$Robotics Institute, Carnegie Mellon University
\quad
$^{2}$Adobe Research
\quad
$^{3}$INRIA \\
\small{\url{http://rohitgirdhar.github.io/ActionVLAD}}
}

\maketitle

\begin{abstract}
In this work, we introduce a new video representation for action classification that
aggregates local convolutional features across the entire spatio-temporal extent of the video. 
We do so by integrating state-of-the-art two-stream networks~\cite{Simonyan_14b}
with learnable spatio-temporal 
feature aggregation~\cite{Arandjelovic16}. 
The resulting architecture is end-to-end trainable for whole-video classification.
We investigate different strategies for pooling across space and time and combining signals from
the different streams. We find that:  (i) it is important to pool jointly across space and time, but (ii) appearance and motion streams are best aggregated into their own separate representations.   
Finally,  we  show  that  our representation  outperforms  the  two-stream  base  architecture by a large margin (13\% relative) as well as
outperforms other baselines with comparable base architectures on
HMDB51, UCF101, and Charades video classification benchmarks.
\end{abstract}

 \section{Introduction}

\begin{figure}
    \centering
    \includegraphics[width=\linewidth]{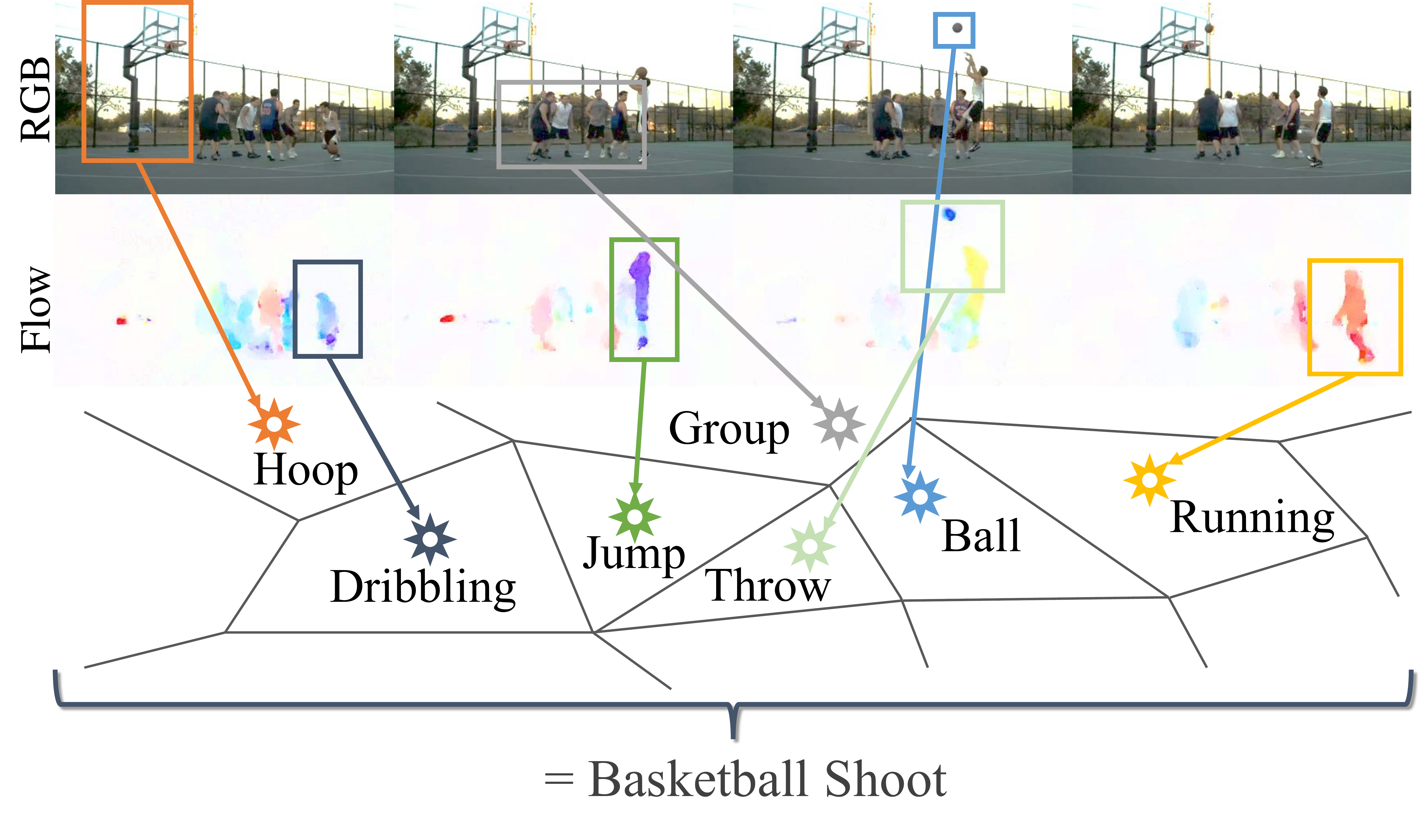}
        \caption{How do we represent actions in a video?
    We propose \methodTag{}, a spatio-temporal aggregation
    of a set of action primitives over the appearance and motion streams
    of a video.
                For example, a basketball shoot may be
    represented as an aggregation of appearance features corresponding to `group
    of players', `ball' and `basketball hoop'; and motion features
    corresponding to `run', `jump', and `shoot'.
    We show examples of primitives our model
    learns to represent videos in Fig.~\ref{fig:cluster_centers}.}
    \label{fig:teaser}
\end{figure}

Human action recognition is one of the fundamental problems in computer vision with applications ranging from video navigation and movie editing to human-robot collaboration.  While there has been great progress in classification of objects in still images using convolutional neural networks
(CNNs)~\cite{He_16,He_16b,Simonyan_14a,Szegedy_16},
this has not been the case for action
recognition.
CNN-based
representations~\cite{Feichtenhofer_16,Varol_16,WangL_16a,WangX_16a,Zhu_15}
have not yet significantly outperformed the best hand-engineered
descriptors~\cite{deSouza_16,IDT_Wang_13}.
This is partly due to missing large-scale video datasets similar in size and variety to ImageNet~\cite{ImageNet}. 
Current video datasets are still rather small~\cite{hmdb51,charades,ucf101} containing only on the order of tens of thousands of videos and a few hundred classes. 
In addition, those classes may be specific to certain domains, such as sports~\cite{ucf101}, 
and the dataset may contain noisy
labels~\cite{Karpathy_14}.
Another key open question is: what is the appropriate {\em spatiotemporal representation} for modeling videos?       
Most recent video representations for action recognition are primarily based on two different CNN architectures: (1) 3D spatio-temporal convolutions~\cite{Tran_15,Varol_16} that potentially learn complicated spatio-temporal dependencies but have been so far hard to scale in terms of recognition performance;  (2) Two-stream architectures~\cite{Simonyan_14b} that  decompose the video into motion and appearance streams, and  train  separate  CNNs  for  each stream,  fusing  the  outputs  in the end.
While both approaches have seen rapid progress,
two-stream architectures have generally outperformed 
the spatio-temporal convolution because
they can easily exploit the new
ultra-deep architectures~\cite{He_16,Szegedy_16}
and models pre-trained for still-image classification.

However, two-stream architectures largely disregard the long-term  temporal structure of the video and essentially learn a classifier that operates on individual frames or short blocks of few (up to 10) frames~\cite{Simonyan_14b}, possibly enforcing consensus of classification scores  over different segments of the video~\cite{WangL_16a}.
At test time, $T$ (typically 25) uniformly  sampled frames (with their motion descriptors) are classified independently and the classification scores are averaged to get the final prediction. 
This approach raises the question whether such temporal averaging is capable of modelling the complex
spatio-temporal structure of human actions. This problem is exaggerated
when the same sub-actions are
shared among multiple action classes.
For example,  consider a complex composite action of a `basketball shoot' shown in Figure~\ref{fig:teaser}.
Given only few consecutive frames of video, it can be easily confused with 
other actions, such as `running', `dribbling', `jumping' and `throwing'.
Using late fusion or averaging is not the optimal solution since it requires frames belonging to same sub-action to be assigned to multiple classes. What we need is a  global feature descriptor for the video which can  aggregate evidence {\em over the entire video} about both the appearance of the scene and the motion of people without requiring every frame to be uniquely assigned to a single action.

To address this issue we develop an end-to-end {\em trainable} video-level representation that aggregates convolutional descriptors across different portions of the imaged scene and across the entire temporal span of the video. At the core of this representation is a spatio-temporal extension of the NetVLAD aggregation layer~\cite{Arandjelovic16} that has been shown to work well 
for instance-level recognition tasks in still images. We call this new layer {\bf \methodTag{}}. 
Extending NetVLAD to video brings the following two main challenges.
First, what is the best way to aggregate frame-level features across time into a video-level representation? To address this, we investigate aggregation at different levels of the network ranging from output probabilities to different layers of convolutional descriptors and show that aggregating the last layers of convolutional descriptors performs best. 
Second, how to best combine the signals from the
different streams in a multi-stream architecture? To address this, we investigate different strategies for aggregating features from spatial and temporal streams and show, somewhat surprisingly, that best results are obtained by aggregating  spatial and temporal streams into their separate single video-level representations. 
We support our investigations with quantitative experimental results together with qualitative visualizations providing the intuition for the obtained results.

 \section{Related Work}

Action recognition is a well studied problem with standard datasets~\cite{youtube8M,2016trecvidawad,THUMOS15,hmdb51,charades,ucf101} focused on tasks
such as classification~\cite{Feichtenhofer_16,Simonyan_14b,IDT_Wang_13,WangL_16a,WangX_16a} and
temporal or spatio-temporal localization~\cite{Kang_16,Weinzaepfel_15,Yeung_15}. 
Action recognition is, however, hard due to the large intra-class variability of different actions and difficulty of annotating large-scale training datasets. As a result, the performance of automatic recognition methods is still far below the capability of human vision. In this paper we focus on the problem of action classification, i.e.,\ classifying a given video clip into one of $K$ given actions classes. We review the main approaches to this problem below followed by a brief review of feature aggregation.

{\bf Dense trajectories:}
Up until recently, the dominating video representation for action recognition has been based on extracting
appearance (such as histograms of image gradients~\cite{Dalal05}) and motion features (such as histogram of flow~\cite{Dalal06})
along densely sampled point trajectories in video. The descriptors are then aggregated into a bag-of-visual-words like representation resulting in a fixed-length descriptor vector for each video~\cite{WangCVPR11,WangBMVC09}. The representation can be further improved by compensating for unwanted camera motions~\cite{IDT_Wang_13}. 
This type of representation, though shallow, is still relevant today and is in fact 
part of the existing state-of-the-art systems~\cite{Feichtenhofer_16,Varol_16,WangL_15a}.
We build on this work by performing a video-level aggregation of descriptors
where both the descriptors and the parameters
for aggregation are jointly learned in
a discriminative fashion.

{\bf Convolutional neural networks:}
Recent work has shown several promising directions in learning video representations directly from data using convolutional neural networks.
For example, Karpathy \etalWspace\cite{Karpathy_14} showed the first large-scale
experiment on training deep convolutional neural networks from a large video
dataset, Sports-1M. Simonyan and Zisserman\cite{Simonyan_14b}
proposed the two-stream architecture, thereby
decomposing a video into appearance and motion information.
Wang \etalWspace\cite{WangL_16a} further improved the two-stream architecture
by
enforcing consensus over predictions in individual frames.
Another line of work has investigated video representations based on spatio-temporal convolutions~\cite{Tran_15,Varol_16}, but these methods have been so far hard to scale to long videos (maximum of 120 frames in \cite{Varol_16}), limiting their ability to learn
over the entire video.

{\bf Modelling long-term temporal structure:}
Some methods explicitly model the temporal structure of the video using, for example, grammars~\cite{Pirsiavash14,Ryoo06} but are often limited to constrained set-ups such as sports~\cite{Msibrahi16}, cooking~\cite{Rohrbach12}, or surveillance~\cite{Amer13}.
More related to our approach, the temporal structure of the video can be also represented implicitly by an appropriate aggregation of descriptors across the video~\cite{Fernando_15,Lev16,Ng_15,Peng14,WangL_15a,Xu_15}. For example, Ng \etalWspace\cite{Ng_15} combine information across frames
using LSTMs.
Xu \etalWspace~\cite{Xu_15} use features from fc7 ImageNet pre-trained
model and use VLAD~\cite{Jegou_10_VLAD} for aggregation, and show
improvements for video retrieval on TRECVID datasets~\cite{2016trecvidawad}.
However, their method is not end-to-end trainable and is used as a post-processing step.
Other works in event detection and action classification rely
on pooling hand crafted features over segments of videos
\cite{Gaidon13,Li13,Niebles2010,Raptis13,Tang12}.
Others have also investigated pooling convolutional descriptors from video
using, for example, variants of Fisher Vectors~\cite{Lev16,Peng14} or pooling
along point trajectories~\cite{WangL_15a}.
In contrast to these methods, we develop an end-to-end trainable video architecture that combines recent advances in two-stream architectures with a trainable spatio-temporal extension of the NetVLAD aggregation layer, which to the best of our knowledge, has not be done before. In addition, we compare favourably the performance of our approach with the above pooling methods in Section~\ref{sec:expts:final}.

{\bf Feature aggregation:}
Our work is also related to feature aggregation such as vectors of locally aggregated descriptors (VLAD)~\cite{Jegou_10_VLAD} and Fisher vectors (FV)~\cite{Perronnin_07,sydorov14cvpr}.
Traditionally, these aggregation techniques have been applied to keypoint descriptors as a post processing
step, and only recently have been extended to  end-to-end training within a convolutional neural network for representing still images~\cite{Arandjelovic16}.
We build on this work and extend it to an end-to-end trainable video representation for action classification by
feature aggregation over space and time.

{\bf Contributions:}
The contribution of this paper are three-fold:
(1) We develop a powerful video-level representation by integrating trainable spatio-temporal aggregation with state-of-the-art two-stream networks. 
(2) We investigate different strategies for pooling across space and time as well as combining signals from the different streams providing insights and experimental evidence for the different design choices. 
(3) We  show  that  our final representation  outperforms  the  two-stream  base  architecture by a large margin (13\% relative) as well as outperforms other baselines with comparable base architectures on
HMDB51, UCF101, and Charades video classification benchmarks.

\begin{figure*}[t]
    \centering
    \includegraphics[width=\linewidth]{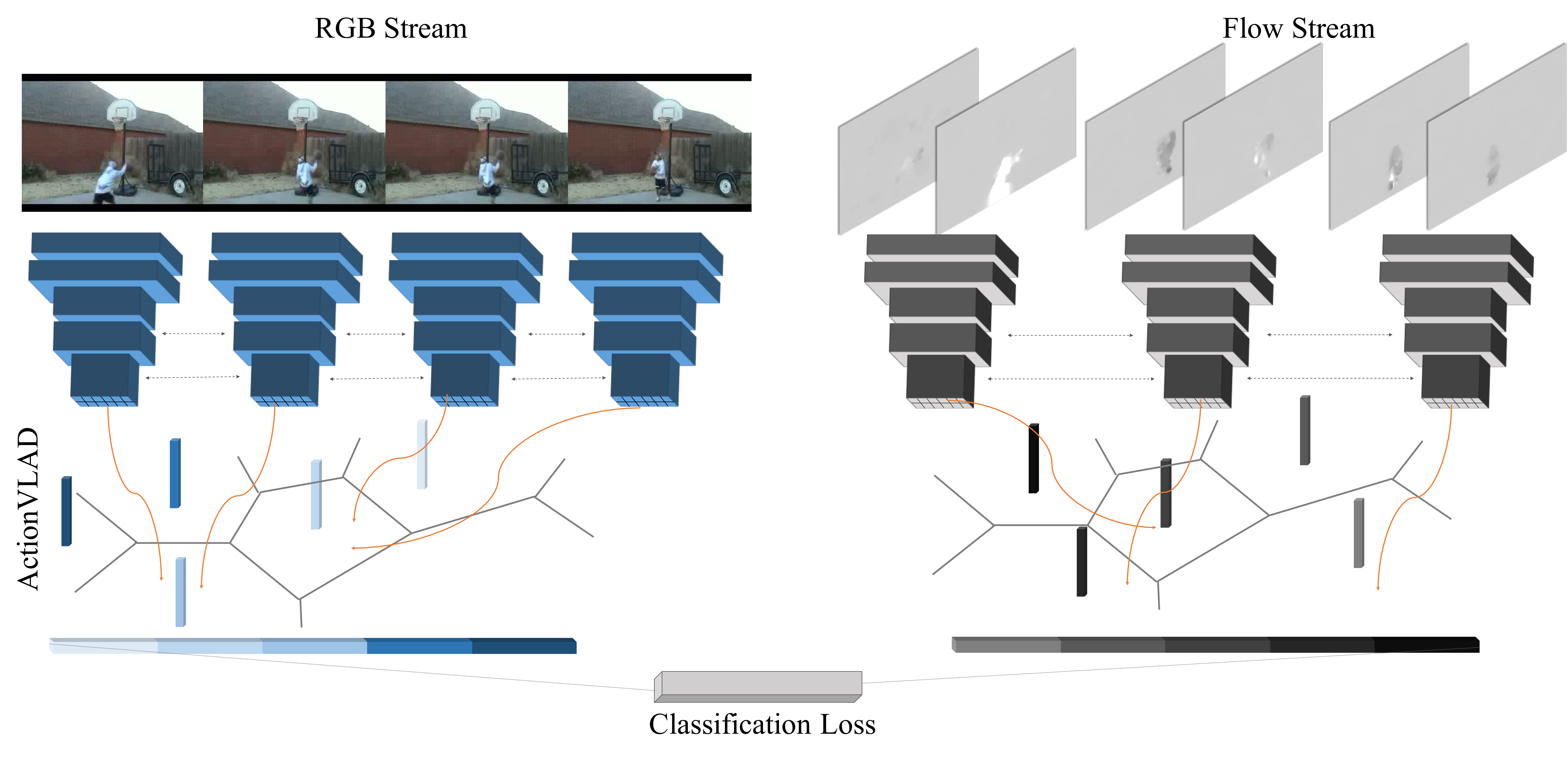}
        \caption{Our network architecture.
    We use a standard CNN architecture (VGG-16)
    to extract features from sampled appearance and motion frames from the video. These
    features are then pooled across space
    and time using the \methodTag{} pooling
    layer, which is trainable end to end with a 
    classification loss. We also experiment with
    \methodTag{} to fuse the two streams (Sec.~\ref{sec:combine-flow-rgb}).}
    \label{fig:nwarch}
\end{figure*}

\section{Video-level two-stream architecture}

We seek to learn a representation for videos that
is trainable end-to-end for action recognition. 
To achieve this we introduce an architecture outlined in Figure~\ref{fig:nwarch}.
In detail, we sample frames from the entire video, and aggregate features from the appearance (RGB)
and motion (flow) streams 
using a vocabulary of ``action words'' into a single video-level fixed-length vector. This representation is then passed through a classifier that outputs the final classification scores. The parameters of the aggregation layer -- the set of ``action words'' -- are learnt together with the feature extractors in a discriminative fashion for the end task of action classification.

In the following we first describe the learnable spatio-temporal aggregation layer (Sec.~\ref{sec:pool-time}). We then discuss the possible placements of the aggregation layer in the overall architecture (Sec.~\ref{sec:pool-where}) and strategies for combining appearance and motion streams (Sec.~\ref{sec:combine-flow-rgb}).
Finally, we give the implementation details (Sec.~\ref{sec:impl-details}).

\subsection{Trainable Spatio-Temporal Aggregation}\label{sec:pool-time}

\begin{figure}[t]
    \centering
    \includegraphics[width=\linewidth]{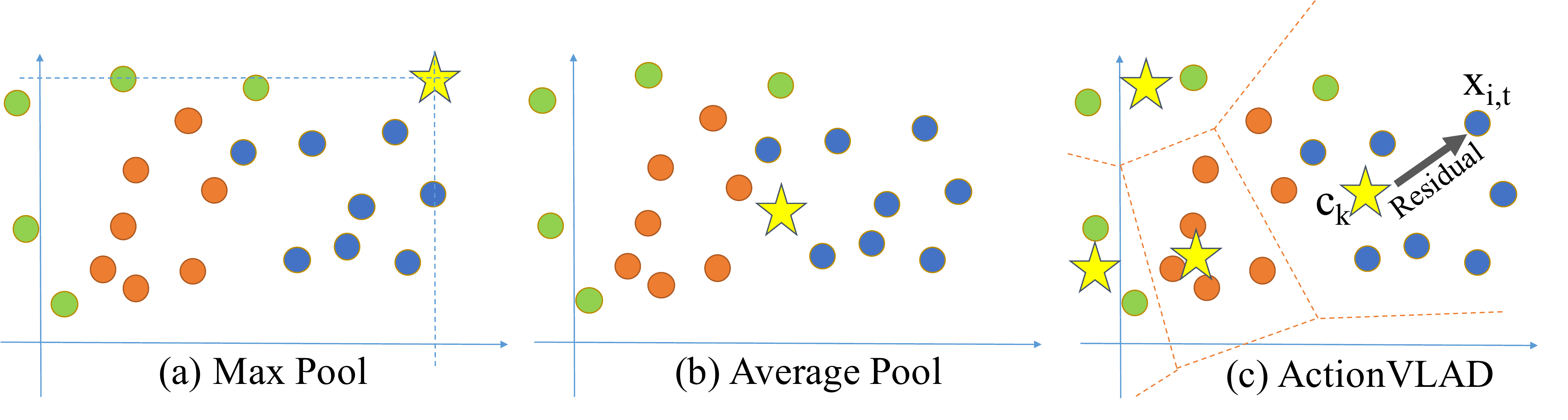}
        \caption{Different pooling strategies for a collection of diverse
    features. 
    Points correspond to features from a video, and 
    different colors correspond to different sub-actions in the video. 
    While (a) max or (b) average pooling are good for similar features,
    they do not not adequately capture the complete distribution of features.
    Our representation (c) clusters the 
    appearance and motion features and aggregates their residuals from nearest cluster centers.}
    \label{fig:pooling-strats}
\end{figure}

Consider $x_{i,t} \in R^D$, a D-dimensional local descriptor extracted from spatial location $i \in \{1 \ldots N\}$ and frame $t \in \{1 \ldots T\}$ of a video. We would like to aggregate these descriptors both spatially and temporally over the whole video while preserving their informative content.
This is achieved by first dividing the descriptor space $R^D$ into $K$ cells using a vocabulary of $K$ ``action words'' represented by anchor points $\{c_k\}$ (Fig.~\ref{fig:pooling-strats} (c)).
Each video descriptor $x_{i,t}$ is then assigned to one of the cells and represented by a residual vector  $x_{it} - c_k$ recording the {\em difference}\ between the descriptor and the anchor point.
The difference vectors are then summed across the entire video as
\begin{equation}
\label{eq:actionVLAD}
        V[j,k] = \sum_{t=1}^{T} \sum_{i=1}^{N}
         \underbrace{\frac{e^{-\alpha||{x_{it}-c_k}||^2}}{\sum_{k'}{e^{-\alpha||{x_{it}- c_{k'}||^2}}}} }_\text{Soft-assignment} 
      \underbrace{\left(
            x_{it}[j] - c_k[j]
        \right)
        }_\text{Residual}
        ,
\end{equation}
where $x_{it}[j]$ and $c_k[j]$ are the $j$-th components of the descriptor vector $x_{it}$ and anchor $c_k$, respectively, and $\alpha$ is a tunable hyper-parameter. Note that the first term in~\eqref{eq:actionVLAD} represents the soft-assignment of descriptor $x_{it}$ to cell $k$ and the second term, $x_{it}[j] - c_k[j]$, is the residual between the descriptor and the anchor point of cell $k$. The two summing operators represent aggregation over time and space, respectively. The output is a matrix $V$, where $k$-th column $V[\cdot,k]$ represents the aggregated descriptor in the $k$-th cell. The columns of the matrix are then intra-normalized~\cite{Arandjelovic13}, stacked, and L2-normalized~\cite{Jegou_10_VLAD} into a single descriptor $v \in R^{KD}$ of the entire video.

The intuition is that the residual vectors record the differences of the extracted descriptors from the ``typical actions" (or sub-actions) represented by anchors $c_k$. The residual vectors are then aggregated across the entire video by computing their sum inside each of the cells. Crucially, all parameters, including the feature extractor, action words $\{c_k\}$, and classifier, are jointly learnt from the data in an end-to-end manner so as to better discriminate target actions. This is because the spatio-temporal aggregation described in~\eqref{eq:actionVLAD} is differentiable and allows for back-propagating error gradients to lower layers of the network. 
Note that the outlined aggregation is a spatio-temporal extension of the NetVLAD~\cite{Arandjelovic16} aggregation, where we, in contrast to~\cite{Arandjelovic16}, introduce the sum across time $t$. We refer to our spatio-temporal extension as {\bf \methodTag{}}.

{\bf Discussion:} It is worth noting the differences of the above aggregation compared to the more common average or max-pooling (Figure~\ref{fig:pooling-strats}). Average or max-pooling represent the entire distribution of points as only a single descriptor which can be sub-optimal for representing an entire video composed of multiple sub-actions. In contrast, the proposed video aggregation represents an entire distribution of descriptors with multiple sub-actions by splitting the descriptor space into cells and pooling inside each of the cells. In theory, a hidden layer between a descriptor map and pooling operation could also split up the descriptor space into half-spaces (by making use of ReLU activations) before pooling. However, it appears difficult to train hidden layers with dimensions comparable to ours $KD=32,768$. We posit that the \methodTag{} framework imposes strong regularization constraints that makes learning of such massive models practical with limited training data (as is the case for action classification).

\subsection{Which layer to aggregate?}\label{sec:pool-where}

In theory, the spatio-temporal aggregation layer described above can be placed at any level of the network to pool the corresponding feature maps. In this section we describe the different possible choices that will later guide our experimental investigation. 
In detail, we build upon the two-stream architecture introduced in
Simonyan and Zisserman~\cite{Simonyan_14b} over a VGG16 network~\cite{Simonyan_14a}. Here we consider only the appearance stream, but discuss the different ways of combining the appearance and motion streams with our aggregation in Section~\ref{sec:combine-flow-rgb}.

Two-stream models first train a frame-level classifier using all the frames
from all videos, and at test time, average the predictions from $T$
uniformly sampled frames~\cite{Simonyan_14b,WangL_16a}. We use
this base network (pre-trained on frame-level) as a feature generator that provides input  from different frames to our trainable \methodTag{} pooling layer.
But, which layer's activations do we pool?
We consider two main choices.
First, we consider pooling the output of fully connected (FC) layers. Those are represented as $1\times 1 $ spatial feature maps with 4096-dimensional output for each of the $T$ frames of the video.  In other words we pool one $4096$-dimensional descriptor from each of the $T$ frames of the video.  Second, we consider pooling features from the convolutional layers (we consider conv4\_3 and conv5\_3). For conv5\_3, for example, those are represented by $14\times 14$ spatial feature maps each with 512-dimensional descriptors for each of the $T$ frames, i.e.\ we pool 196 512-dimensional descriptors from each of the $T$ frames. 
As we show in Section~\ref{sec:expts:whereNetVLAD}, we obtain 
best performance by pooling features at the highest convolutional layer
(conv5\_3 for VGG-16).

\subsection{How to combine Flow and RGB streams?}\label{sec:combine-flow-rgb}

\begin{figure}
    \centering
    \begin{subfigure}{0.32\linewidth} \centering
     \includegraphics[width=\linewidth]{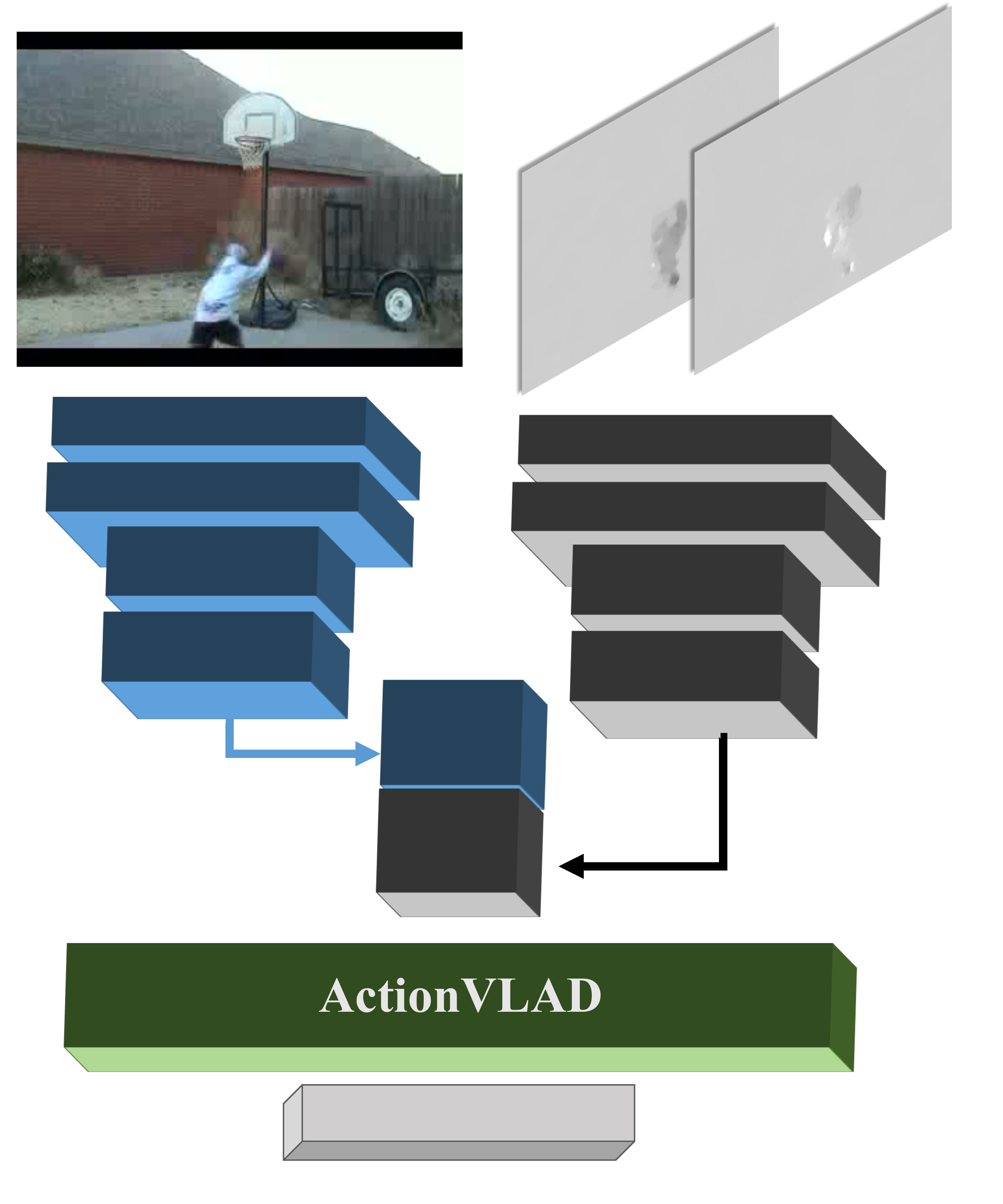}
     \caption{Concat Fusion}\label{fig:rgbflow:concat}
   \end{subfigure}
   \begin{subfigure}{0.32\linewidth} \centering
     \includegraphics[width=\linewidth]{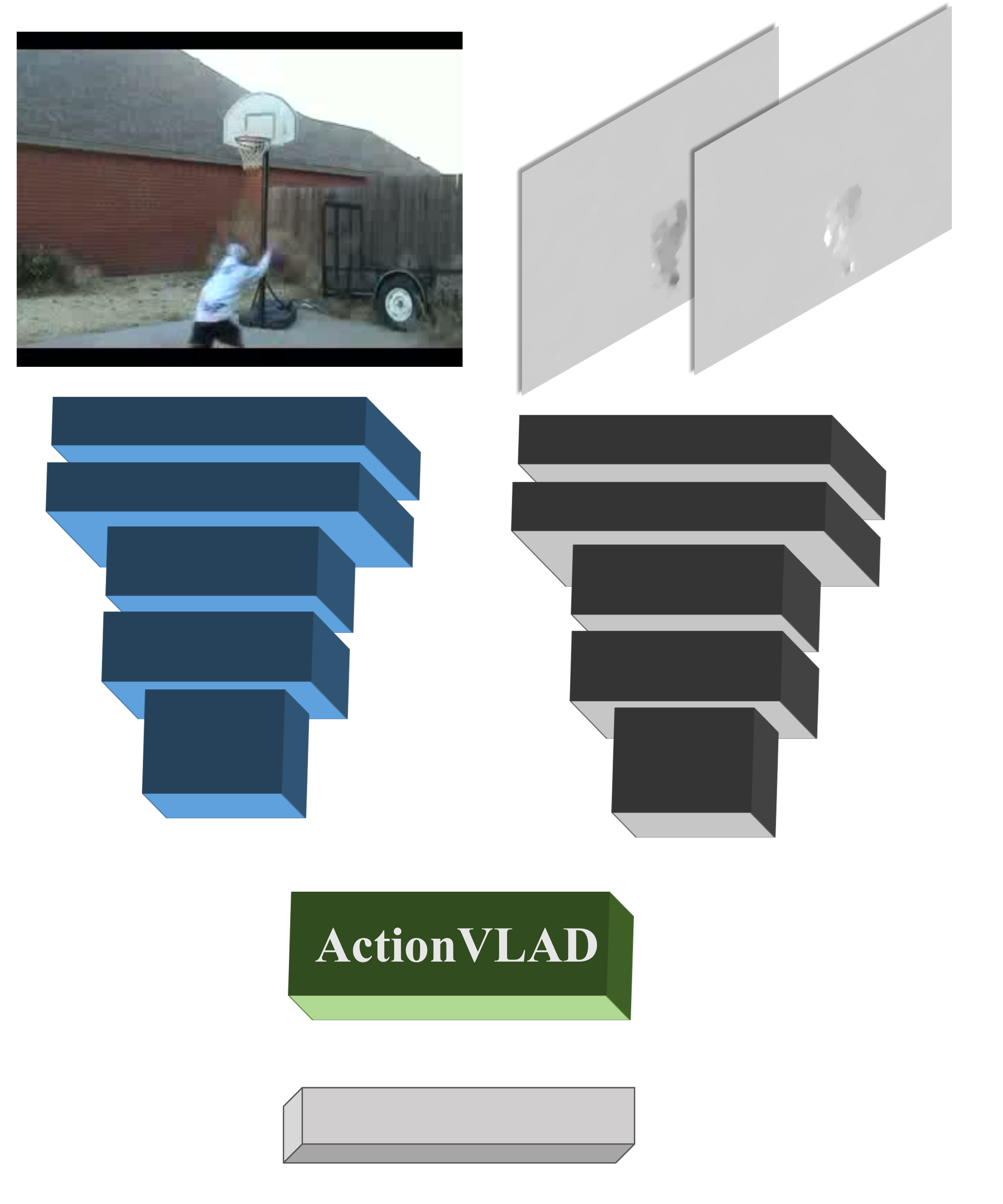}
     \caption{Early Fusion}\label{fig:rgbflow:onebag}
   \end{subfigure}
   \begin{subfigure}{0.32\linewidth} \centering
     \includegraphics[width=\linewidth]{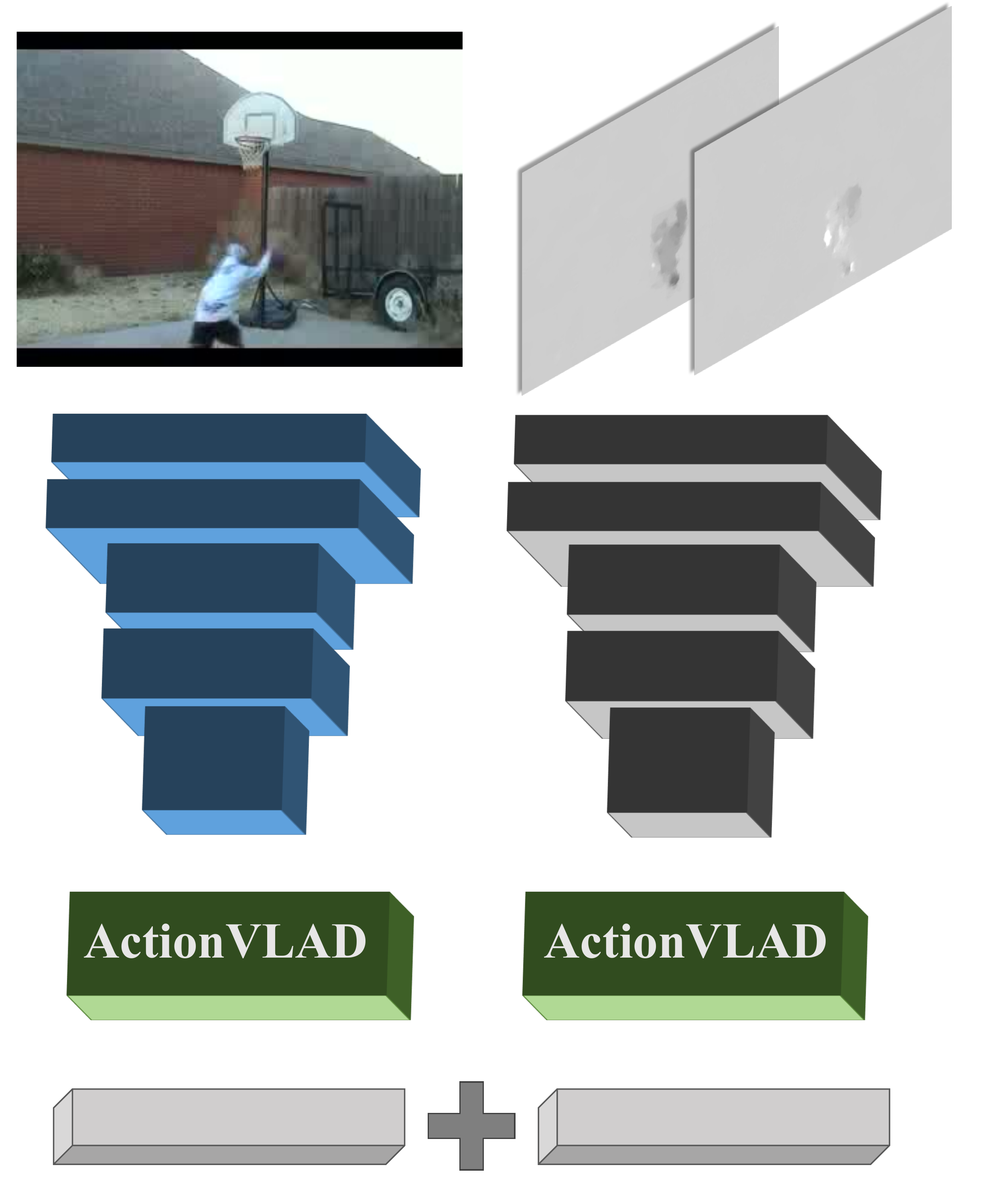}
     \caption{Late Fusion}\label{fig:rgbflow:late}
   \end{subfigure}
    \caption{Different strategies for combining the appearance and motion streams.
        }\label{fig:rgbflow-pooling}
\end{figure}

\methodTag{} can be also used to pool features across the different streams of input
modalities. In our case we consider appearance and motion streams~\cite{Simonyan_14b}, but in theory pooling can be done across any number of other data streams,
such as  warped flow or RGB differences~\cite{WangL_16a}.
There are several possible ways to combine the appearance and motion streams to obtain a
jointly trainable representation. We explore the most salient ones in this section and outline them in Figure~\ref{fig:rgbflow-pooling}.

{\noindent \bf Single \methodTag{} layer over concatenated appearance and motion features (Concat Fusion).}
In this case, we concatenate the corresponding output feature maps from appearance and motion, essentially assuming their spatial correspondence. We place a single \methodTag{} layer on top of this concatenated feature map, as shown in Figure~\ref{fig:rgbflow-pooling}(a). This allows correlations between appearance and flow features to be exploited for codebook construction.

{\noindent \bf Single \methodTag{} layer over all appearance and motion features (Early Fusion).}
We also experiment with pooling all features from appearance and motion streams
using a single \methodTag{} layer, as illustrated in Figure~\ref{fig:rgbflow-pooling}(b). This encourages the model to learn a single descriptor space $x_{ij}$ for both appearance and motion features,
and hence exploit redundancy in the features.

{\noindent \bf Late Fusion.}
This strategy, shown Figure~\ref{fig:rgbflow-pooling}(c), follows the standard testing practice of weighted average of the appearance and motion
last layer features.
Hence, we have 
two separate \methodTag{} layers, one for each stream.
This allows for both \methodTag{} layers to learn specialized representations
for each input modality.

We compare the performance of the different aggregation techniques in Section~\ref{sec:expts:flowRGB}.

\subsection{Implementation Details}\label{sec:impl-details}

We train all our networks with a single-layer linear
classifier on top of our \methodTag{} representation
described above. Throughout, we use $K=64$
and a high value for $\alpha=1000.0$, similar to~\cite{Arandjelovic16}.
Since the output feature dimensionality can be
large, we use a dropout of 0.5 over the 
representation to avoid overfitting to small action classification
datasets. We train the network with cross-entropy loss, where 
the probabilities are obtained through a softmax. Similar
to~\cite{Arandjelovic16}, we decouple \methodTag{} parameters $\{c_k\}$ used to compute the soft assignment and the residual from \eqref{eq:actionVLAD} to simplify learning (though both sets of parameters are identically initialized with the same cluster centers).

We use $T=25$ frames per video (for both flow and RGB) to learn and evaluate our video representation,
in order to be comparable to standard practice~\cite{Simonyan_14b,WangL_16a}.
Flow is represented using 10 consecutive $x$ and $y$ direction flow maps, to get 
a 20-channel input.
Since our model is trained at a video level, we can fit very few videos every iteration
due to limited GPU memory and CPU preprocessing capability. To maintain reasonable batch sizes, we use slow updates by 
averaging gradients over multiple GPU iterations.
We clip gradients at 5.0 L2 norm.
Data augmentation is done at the video level by 
performing random cropping/flipping of all the RGB and flow
frames correspondingly.
When training \methodTag{}, we use the Adam solver~\cite{Kingma14} with
$\epsilon=10^{-4}$. This is required as the \methodTag{} output is L2 normalized
and we need a lower $\epsilon$ value for reasonably fast convergence.
We perform training in a two-step approach. In
the first step, we initialize and fix the VLAD cluster centers,
and only train the linear softmax
classifier with a learning rate of 0.01.
In the second step, we jointly finetune both
the linear classifier and the \methodTag{}
cluster centers with a learning 
rate of $10^{-4}$.
Our experiments show that this uniformly gives a significant boost in validation accuracy (Table~\ref{tab:train_thru_netvlad}), indicating that \methodTag{} does adapt clusters to better represent the videos.
When training \methodTag{} over conv5\_3,
we keep layers before conv5\_1 fixed
to avoid overfitting to small action classification datasets.
This also helps by having a smaller GPU memory footprint and faster training.
That said, our model is completely capable of end-to-end training,
which can be exploited for larger, more complex datasets.
We implemented our models in TensorFlow~\cite{tensorflow2015-whitepaper} and have released our code and pretrained models~\cite{webpage}.

 \section{Experiments}

In this section we experiment with the various network
architectures proposed above on standard
action classification benchmarks.

\subsection{Datasets and Evaluation}
We evaluate our models on two popular trimmed action
classification benchmarks,
UCF101~\cite{ucf101} and HMDB51~\cite{hmdb51}.
UCF101 consists of 13320 sports video clips
from 101 action classes,
and HMDB51 contains 6766 realistic and varied video clips from
51 action classes. We follow the evaluation
scheme from THUMOS13 challenge~\cite{THUMOS13} and
employ the provided three train/test splits for
evaluation.
We use the split 1 for ablative analysis and
report final performance averaged over all 3 splits.
Finally, we also evaluate our model on an
untrimmed dataset, Charades~\cite{charades}.
Since a video in Charades can have multiple labels,
the evaluation is performed using mAP and weighted
average precision (wAP), where AP of each class
is weighted by class size.

\subsection{Training \methodTag{} Representation}

To first motivate the strength of the trainable \methodTag{} representation,
we compare it to non-trained \methodTag{} extracted from video.
In particular, we train only a single classifier over the \methodTag{} layer (over conv5\_3) initialized by k-means (and kept fixed).
Next, we start from the above model 
and also train the parameters of the 
layers including and before \methodTag{},
upto conv5\_1. As shown in Table~\ref{tab:train_thru_netvlad}, 
removing the last 2-layer non-linear classifier from the
two-stream model and training a single 
linear layer over the (fixed) VLAD pooled
descriptor already gets close to the two-stream performance.
This improves even further when the parameters of \methodTag{} are trained together with the preceding layers.
Throughout, we use the default value of $K=64$
following~\cite{Arandjelovic16}. Initial 
experiments (HMDB51 split 1 RGB) have shown stable 
performance on varying $K$
($49.1\%$, $51.2\%$ and $51.1\%$ for $K=$
32, 64 and 128 respectively).
Reducing $K$ to 1, however, leads
to much lower performance, $43.2\%$.

\begin{table}[]
\caption{Evaluation of training \methodTag{} representation
using VGG-16 architecture on HMDB51 split 1.}
\label{tab:train_thru_netvlad}
\tableSize{}
\centering
\begin{tabular}{llll}
\toprule
Method    & Appearance  & Motion      &  \\
\midrule
Two Stream~\cite{Feichtenhofer_16} & 47.1 & 55.2 \\
Non-trained \methodTag{} + Linear Classifier & 44.9 & 55.6 \\
Trained \methodTag{} & 51.2 & 58.4 \\
\bottomrule
\end{tabular}
\end{table}

We also visualize the classes for which we obtained the highest improvements compared to the standard two-stream model.
We see the largest gains for classes such as `punch', which
were often confused with `kick' or `hit';
`climb\_stairs', often confused with `pick' or `walk';
`hit', often confused with `golf'; and `drink', 
often confused with `eat' or `kiss'. This is expected,
as these classes are easy to confuse given only local visual evidence.
For example, hitting and punching are easily confused when looking at few frames but can be disambiguated when aggregating information over the whole video. We present the whole confusion matrix highlighting the pairs of classes with the highest changes of performance in the appendix~\cite{appendix}.

\subsection{Where to \methodTag{}?}\label{sec:expts:whereNetVLAD}

Here we evaluate the different positions in the network where we
can insert the \methodTag{} layer. Specifically, we compare placing \methodTag{}  after the last two 
convolutional layers (conv4\_3 and conv5\_3) and after the last fully-connected layer (fc7). In each case, we train until the block of layers just before
the \methodTag{} layer; so conv4\_1 to loss in case of conv4, and 
fc7 to loss in case of fc7.
Results are shown in Table~\ref{tab:ablative-fc-conv} and clearly show that best performance is obtained by aggregating the last convolutional layer (conv5\_3). fc6 features obtain performance similar to fc7,
getting $42.7\%$ for RGB.

\begin{figure}
    \centering
    \begin{subfigure}{0.49\linewidth} \centering
     \includegraphics[width=\linewidth]{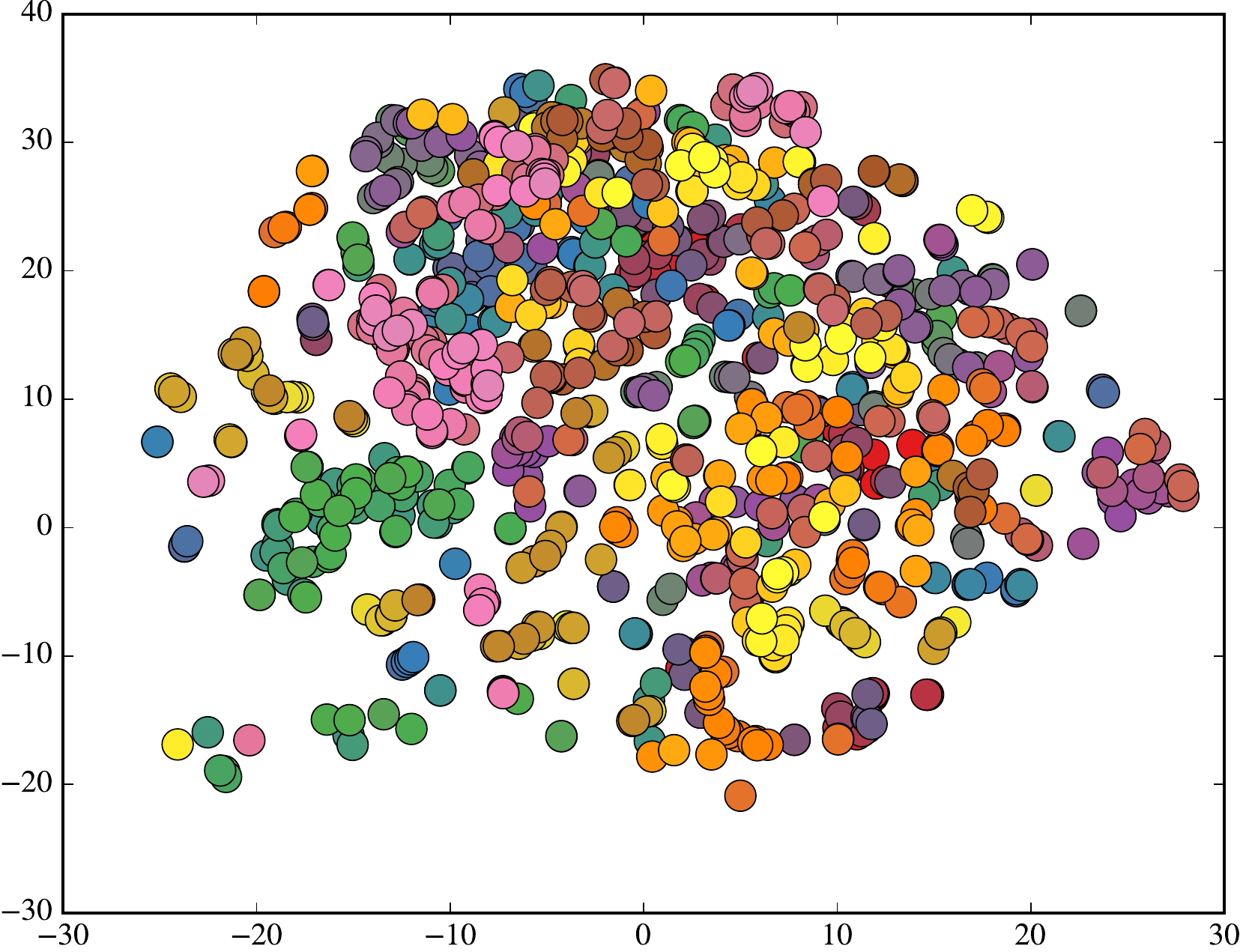}
     \caption*{conv5\_3}\label{fig:rgbflow:concat}
   \end{subfigure}\hfill
   \begin{subfigure}{0.49\linewidth} \centering
     \includegraphics[width=\linewidth]{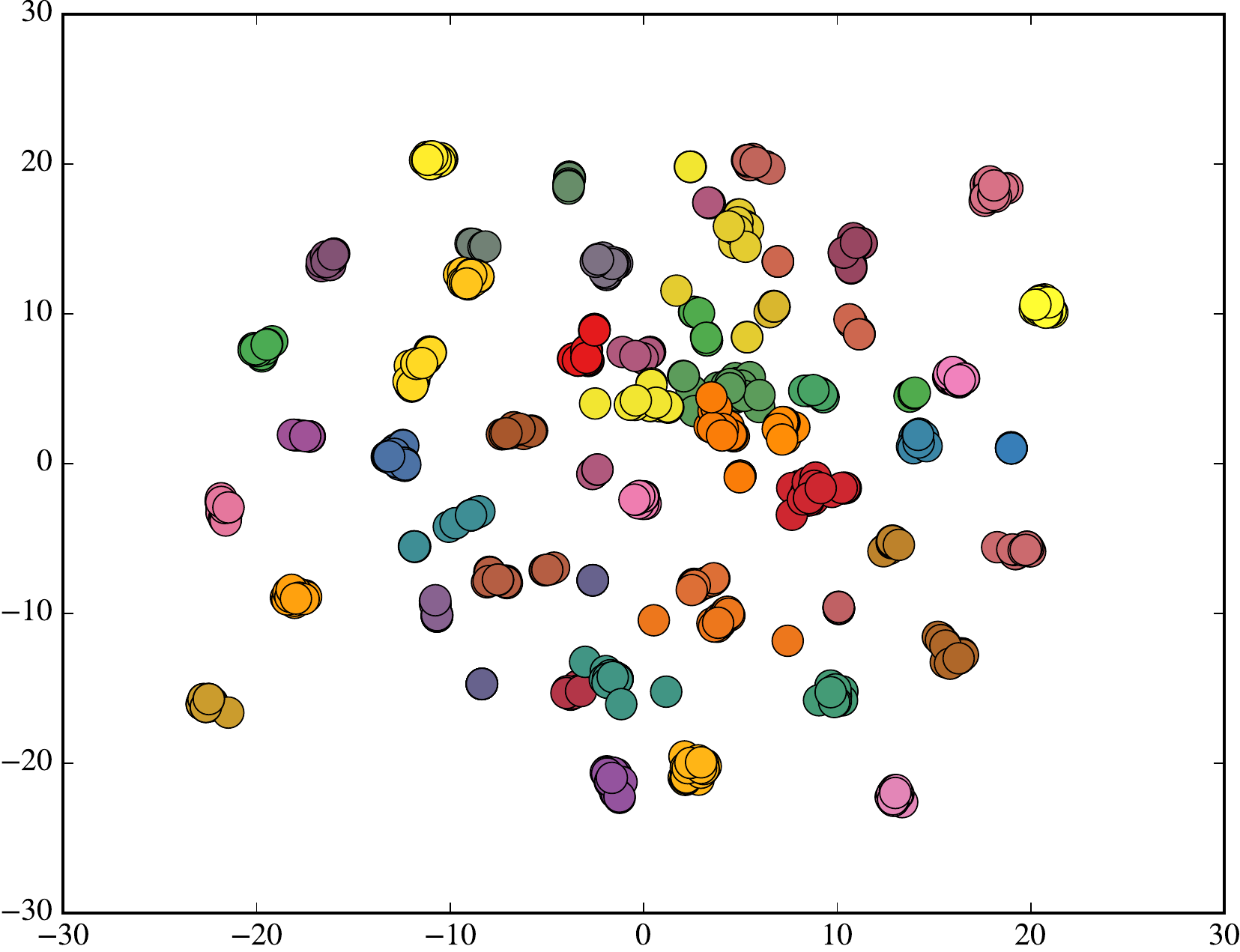}
     \caption*{fc7}\label{fig:rgbflow:concat}
   \end{subfigure}
       \caption{tSNE embedding of subsampled conv and fc features from a set of videos.
    Features belonging to the same video have the same color.
        Note that the fc features are already very similar and hence cluster together,
    while the conv5\_3 features are more spread out and mixed.}
    \label{fig:tsne_fc_conv5}
\end{figure}

We believe that this is due to two reasons. First, pooling
at a fully-connected layer prevents \methodTag{} 
from modelling spatial information, as these layers
already compress a lot of information.
Second, fc7 features are more semantic and hence features from different frames are already  similar to each other not taking advantage 
of the modelling capacity of the  \methodTag{} layer, i.e.\ they would often fall into the same cell.
To verify this hypothesis we visualize the conv5\_3 and fc7 appearance features from the same frames using a tSNE~\cite{tsne} embedding in Figure~\ref{fig:tsne_fc_conv5}.
The plots clearly show that fc7 features from the same video (shown in the same color) are already
similar to each other, while conv5\_3 features are much more varied and ready to benefit from the ability of the \methodTag{} layer to capture complex distributions in the feature space as discussed in Sec.~\ref{sec:pool-time}.

\subsection{Baseline aggregation techniques}\label{sec:expts:whichAgg}

Next, we compare our ActionVLAD aggregation with baseline aggregation techniques. As discussed earlier, average or max pooling reduce the distribution of features in the video into one single point in the feature space, which is suboptimal for the entire video.  This is supported by our results in Table~\ref{tab:ablative-max-avg-netvlad}, where we see a significant drop in performance using average/max pooling over con5\_3 features, even compared to the baseline two-stream architecture.

\begin{table}
    \caption{Evaluation of (a) \methodTag{} at different positions in a VGG-16 network;
    and (b) \methodTag{} compared to other pooling strategies, on HMDB51 split 1.}
    \begin{subfigure}{0.49\linewidth}
        \centering
        \caption{Position of \methodTag{}.}       
        \label{tab:ablative-fc-conv}
        \tableSize{}
        \centering
        \begin{tabular}{lll}
        \toprule
        Method               & RGB      & Flow \\ \midrule
        2-Stream           & 47.1   & 55.2 \\
        conv4\_3        & 45.0 &  53.5   \\
        conv5\_3       & 51.2 & 58.4 \\
        fc7         & 43.3 & 53.1 \\ \bottomrule
        \end{tabular}
    \end{subfigure}\hfill
    \begin{subfigure}{0.49\linewidth}
        \caption{Different pooling strategies.}\label{tab:ablative-max-avg-netvlad}
        \tableSize{}
        \centering
        \begin{tabular}{lll}
        \toprule
        Method               & RGB      & Flow \\ \midrule
        2-Stream           & 47.1   & 55.2 \\
        Avg        & 41.6 &  53.4   \\
        Max         & 41.5 & 54.6 \\
        \methodTag{} & 51.2 & 58.4 \\ \bottomrule
        \end{tabular}
    \end{subfigure}
\end{table}

\subsection{Combining Motion and Appearance}\label{sec:expts:flowRGB}

We compare the different combination strategies in Table~\ref{tab:ablative-flow-rgb}.
We observe that late fusion performs best.
To further verify this observation, we show a tSNE plot of 
conv5 features from appearance and motion streams in the appendix~\cite{appendix}.
The two features are well separated, indicating there is potentially
complementary information that is best exploited by fusing
later in the network.
In contrast, concat fusion limits the modelling power of the model as it uses the same number of cells to capture a larger portion of the feature space. Finally, we compare our overall performance
on three splits of HMDB51 with the two-stream baseline in Table~\ref{tab:overall_vs_2stream}.
We see significant improvements over each input modality as well as over the
final (late-fused) vector.

\begin{table}
    \caption{Comparison of (a) Different fusion techniques described
    in Sec.~\ref{sec:combine-flow-rgb} on HMDB split 1;
    and (b) Comparison of two-stream with \methodTag{},
    averaged over 3-splits of HMDB.}
    \begin{subfigure}[t]{0.4\linewidth}
        \centering
        \caption{Fusion Type}       
        \label{tab:ablative-flow-rgb}
        \tableSize{}
        \centering
        \begin{tabular}{ll}
        \toprule
        Method               & Val acc \\ \midrule
                Concat       & 56.0 \\
        Early         & 64.8 \\
        Late     & 66.9 \\ \bottomrule
                        \end{tabular}
    \end{subfigure}\hfill
    \begin{subfigure}[t]{0.6\linewidth}
        \caption{Overall Comparison}\label{tab:overall_vs_2stream}
        \tableSize{}
        \centering
        \begin{tabular}{lll}
        \toprule
        Stream               & 2-Stream~\cite{WangX_16a} & Ours \\ \midrule
        RGB       & 42.2   & 49.8 \\
        Flow      & 55.0   & 59.1 \\
        Late Fuse & 58.5   & 66.3 \\ \bottomrule
        \end{tabular}
    \end{subfigure}
\end{table}

\begin{figure*}[t]
    \centering
    \includegraphics[width=\linewidth]{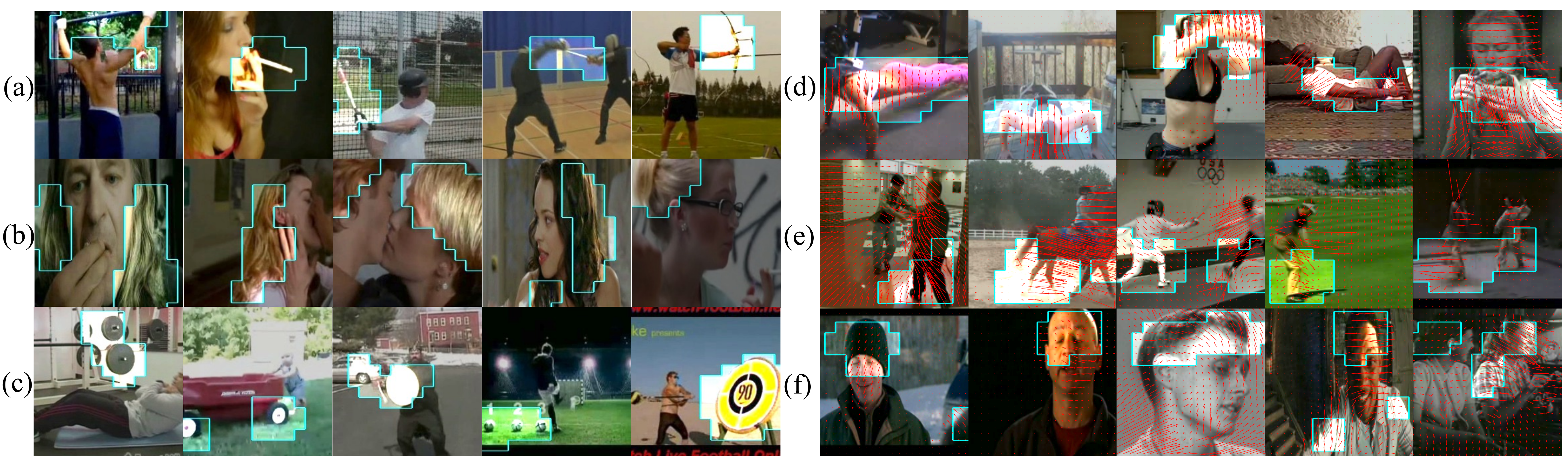}\hfill
            \caption{Visualization of `action words' our \methodTag{} model learns when trained
    for appearance and motion modalities.
    Each row shows several frames from videos where bright regions correspond to the center of 
    the receptive field for conv5\_3 features that get assigned to one specific `action word' cell.
    In detail, (a) shows an `action word' that looks for human hands holding rod-like objects,
    such as a pull-up bar, baseball bat or archery bow. (b) looks for human hair.
    (c) looks for circular objects like wheels and target boards.
    (d)-(f) shows similar action words for the flow stream. These are more complex, as each
    word looks at shape and motion of the flow field over 10 frames. Here we show some easy to interpret cases,
    such as (d) up-down motion, (e) linear motion for legs and (f) head motion.}
    \label{fig:cluster_centers}
    \end{figure*}

\begin{figure}[t]
    \centering
    \includegraphics[width=\linewidth]{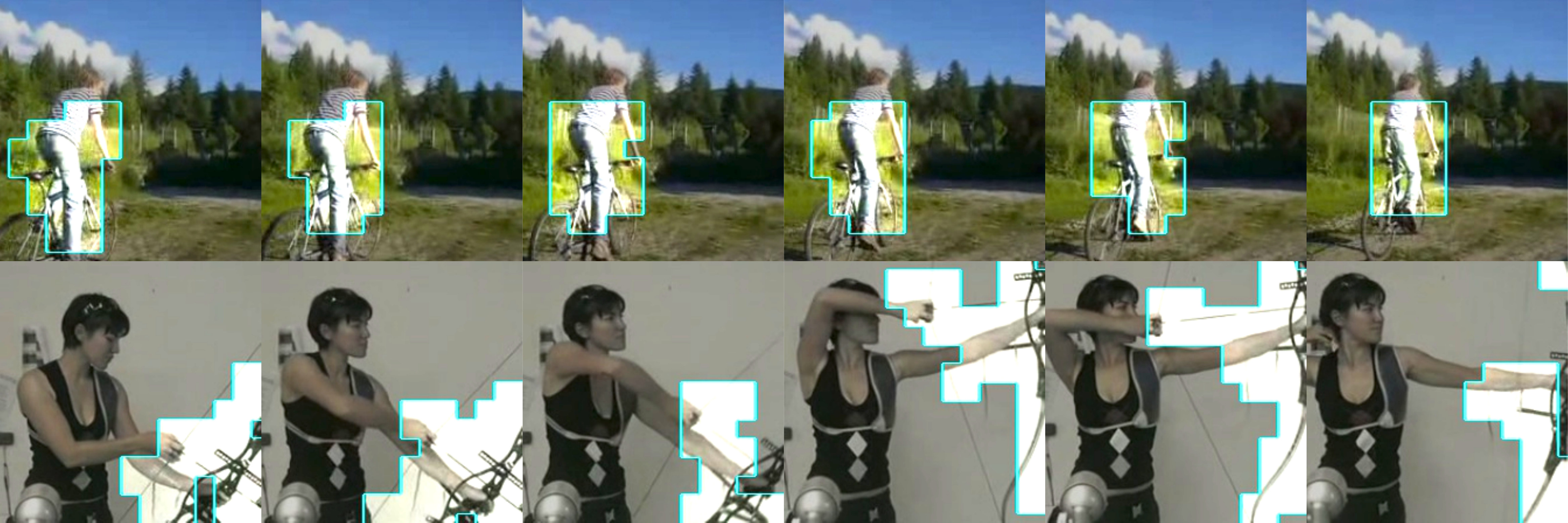}
    \caption{Assignment of image regions to two specific appearance `action words' over several video frames. \methodTag{} representation ``tracks''
    these visual and flow patches over the video.
            }\label{fig:cluster_center_movement}
\end{figure}

\begin{figure}[t]
    \centering
    \includegraphics[width=\linewidth]{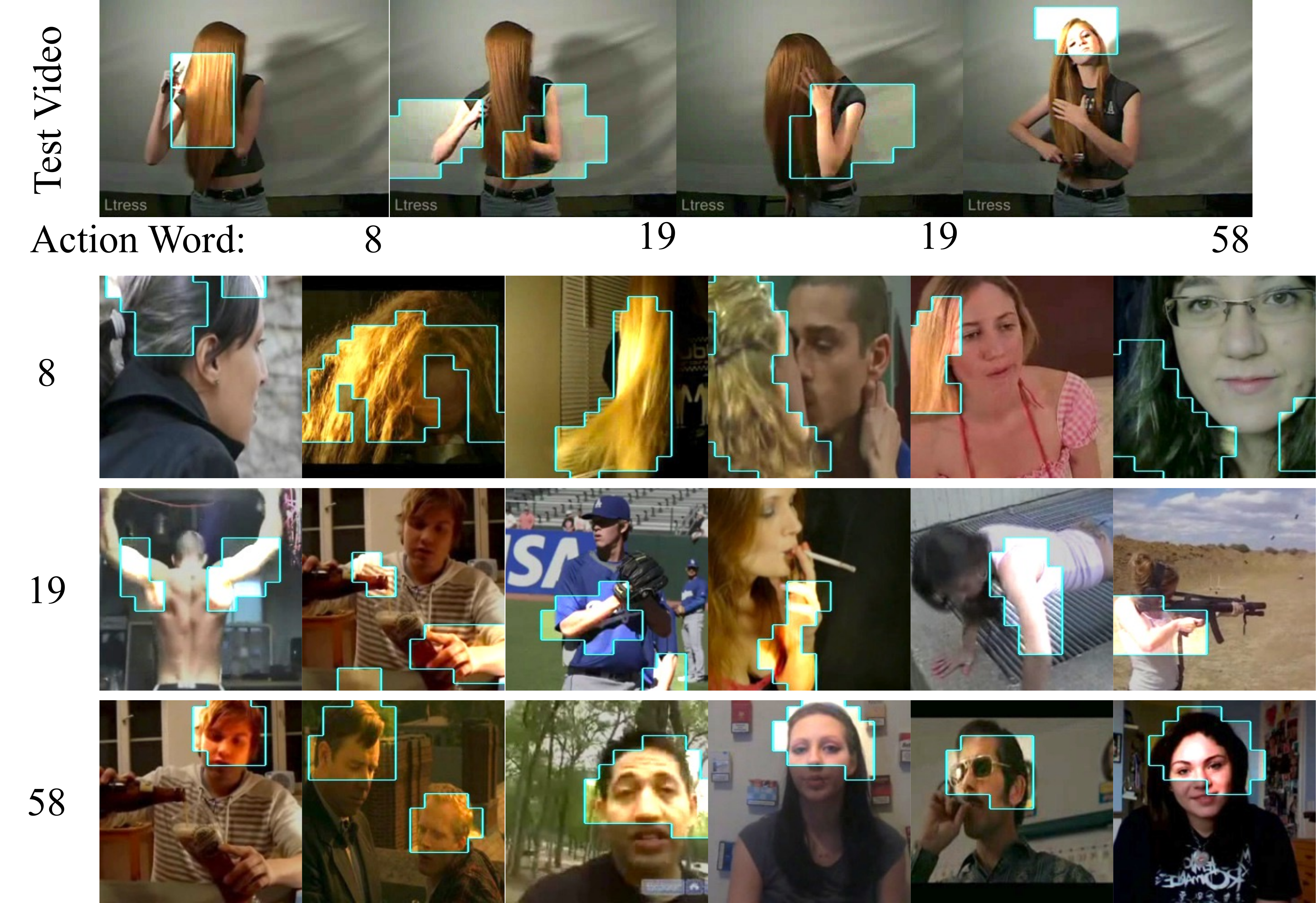}
        \caption{
    Top contributing action words for classifying this video.  
    The softmax score for this `brushing
    hair' video is obtained (in appearance stream) from features corresponding to `action words'
    that seem to focus on
    faces, eyes, hair and hand/leg clusters. It was incorrectly classified as `clap' by the baseline two-stream architecture but is correctly classified as `brushing hair' by our method.}\label{fig:qual_res}
  \end{figure}

\subsection{Comparison against State of the Art}\label{sec:expts:final}

\begin{table}[]
\centering
\caption{Comparison with the state of the art
on UCF101 and HMDB51 datasets averaged 
over 3 splits. 
First section compares all ConvNet-based methods reported
using VGG-16 or comparable models.
Second section compares methods that use iDT~\cite{IDT_Wang_13},
and third section reports methods using ultra deep architectures,
multi-modal inputs (more than RGB+Flow) and 
hybrid methods.
}
\label{tab:main-compare}
\tableSize{}
\begin{tabular}{lll}
\toprule
                                        & UCF101 & HMDB51 \\ \midrule
Spatio-Temporal ConvNet~\cite{Karpathy_14} & 65.4 & - \\
LRCN~\cite{LRCN} & 82.9 & - \\
C3D~\cite{Tran_15}                             & 85.2 & -      \\
Factorized ConvNet~\cite{Sun15}              & 88.1 & 59.1 \\
VideoDarwin~\cite{Fernando_15} & - & 63.7 \\
Two-Stream + LSTM~\cite{Ng_15} (GoogLeNet)   & 88.6 & -      \\
Two-Stream ConvNet~\cite{Simonyan_14b} (VGG-M)      & 88.0 & 59.4 \\
Two-Stream ConvNet~\cite{WangL_15_GoodPrac,WangX_16a} (VGG-16)     & 91.4 & 58.5 \\
Two-Stream Fusion~\cite{Feichtenhofer_16} (VGG-16)	        & 92.5  & 65.4 \\
TDD+FV~\cite{WangL_15a}  & 90.3 & 63.2 \\
RNN+FV~\cite{Lev16} & 88.0 & 54.3 \\
Transformations~\cite{WangX_16a} & 92.4 & 62.0 \\
LTC~\cite{Varol_16} & 91.7 & 64.8 \\
KVMF~\cite{Zhu_15} & {\bf 93.1} & 63.3 \\
{\bf \methodTag{} (LateFuse, VGG-16)}                  & 92.7  & {\bf 66.9} \\ 
\midrule
DT+MVSV~\cite{Cai_14} &  83.5 & 55.9 \\
iDT+FV~\cite{IDT_Wang_13} & 85.9 & 57.2 \\
iDT+HSV~\cite{Peng_16} & 87.9 & 61.1 \\
MoFAP~\cite{WangL_16b} & 88.3 & 61.7 \\
  C3D+iDT~\cite{Tran_15} & 90.4 & - \\
Two-Stream Fusion+iDT~\cite{Feichtenhofer_16}        & 93.5  & 69.2 \\
LTC+iDT~\cite{Varol_16} & 92.7 & 67.2 \\
{\bf \methodTag{} (VGG-16) + iDT} & {\bf 93.6} & {\bf 69.8} \\
\midrule
TSN (BN-Inception, 3-modality)~\cite{WangL_16a}  &  94.2 & 69.4 \\
DT+Hybrid architectures~\cite{deSouza_16} & 92.5  & 70.4 \\
ST-ResNet+iDT~\cite{Feichtenhofer_16b} & 94.6 & 70.3 \\
\bottomrule
\end{tabular}
\end{table}

\begin{table}[]
\centering
\caption{Comparison with the state of the art
on Charades~\cite{charades}
using mAP and weighted-AP (wAP)
metrics.}
\label{tab:charades-compare}
\tableSize{}
\begin{tabular}{lll}
\toprule
                                        & mAP & wAP \\ \midrule
Two-stream + iDT (best reported)~\cite{charades} & 18.6 & - \\
RGB stream (BN-inception, TSN~\cite{WangL_16b} style training) & 16.8  & 23.1 \\
ActionVLAD (RGB only, BN-inception) & 17.6 & 25.1 \\
ActionVLAD (RGB only, BN-inception) + iDT & {\bf 21.0} & {\bf 29.9} \\
\bottomrule
\end{tabular}
\end{table}

In Table~\ref{tab:main-compare}, we compare our approach to a variety of recent action recognition methods that use a comparable base architecture to ours (VGG-16). Our model outperforms all previous approaches on HMDB51 and UCF101,
using comparable base architectures when combined with iDT.
Note that similar to 10-crop testing in two-stream models, our model
is also capable of pooling
features from multiple crops at test time.
We report our final performance using
5 crops, or 125 total images, per video.
Other methods such as~\cite{Feichtenhofer_16b}
and~\cite{WangL_16a} based on ultra-deep architectures such as ResNet~\cite{He_16} and
Inception~\cite{Ioffe_15} respectively obtain
higher performance. However, it is interesting to
note that our model still outperforms some of these ultra-deep two-stream baselines,
such as
ResNet-50 (61.2\% on HMDB51 and 91.7\% on UCF101) and
    ResNet-152 (63.8\% and 91.8\%)~\cite{Feichtenhofer_16b}\footnote{{\tiny Reported on \url{http://www.robots.ox.ac.uk/~vgg/software/two_stream_action/}}},
while employing only a VGG-16 network with ActionVLAD (66.9\% and 92.7\%).
We evaluate our method on
Charades~\cite{charades} in Tab.~\ref{tab:charades-compare},
and outperform all previous reported methods (details in appendix~\cite{appendix}).

\subsection{Qualitative Analysis}

Finally, we visualize what our model has learned.
We first visualize the learned `action words', to understand the primitives
our model uses to represent actions. We randomly picked few thousand frames from
HMDB videos, and computed \methodTag{} assignment maps for conv5\_3
features by taking the max 
over soft-assignment in Eq.~\ref{eq:actionVLAD}.
These maps define which features 
get soft-assigned to which of the 64 `action words'.
Then for each action word, we visualize the
frames that contain that action word, and highlight 
regions that correspond to center of 
receptive field of conv5\_3 neuron assigned to that word.
Fig.~\ref{fig:cluster_centers} shows some such action
words and our interpretation for those words.
We visualize 
these assignment maps 
over videos in Fig.~\ref{fig:cluster_center_movement}.

To verify how the `action words' help in classification, we consider 
an example video in Fig.~\ref{fig:qual_res} that was originally misclassified
by the two-stream model but gets a high softmax score for the correct class with our
\methodTag{}
model. For ease of visualization, we only consider the RGB model in this example. 
We first compute the \methodTag{} feature for this video, extract the linear
classifier weights for the correct class, and accumulate the contribution
of each `action word' in the final score. We show a visualization for some
of the distinct top contributing words. The visualization for each word follows same
format as in Fig.~\ref{fig:cluster_centers}. We see that this `brushing hair' video
is represented by spatial `action words' 
focussing on faces, eyes, hair, and hands.
 \section{Conclusion}

We have developed a successful approach for spatiotemporal video feature aggregation for action classification. Our method is end-to-end trainable and outperforms most prior approaches based on the two-stream VGG-16 architecture on the HMDB51 and UCF101 datasets. Our approach is general and may be applied to future video architectures as an ActionVLAD CNN layer, which may prove helpful for related tasks such as (spatio) temporal localization of human actions in long videos.

 {\footnotesize
\paragraph{Acknowledgements:}
The authors would like to thank G{\"u}l Varol and Gunnar
Atli Sigurdsson for help with iDT.
DR is supported by NSF Grant 1618903, Google, and the Intel Science and Technology Center for Visual Cloud Systems (ISTC-VCS).
}
 
{\small
\bibliographystyle{ieee}
\bibliography{refs}
}

\clearpage
\appendix

\section*{Appendices}

\section{Multi-crop Testing}
Similar to the two-stream approaches~\cite{WangL_15_GoodPrac,WangL_16a} that typically use multiple crops and flips of the video frames at test time to average predictions over, we can also perform \methodTag{} pooling over multiple crops of the input frames.
All the performance evaluation of \methodTag{} in the main paper
uses a single center crop from each frame at test time (single-crop testing).
In Table~\ref{tab:multi-crop}, we 
compare the single-crop testing with 
a multi-crop strategy. Specifically, we use 4 corner
crops along with the center crop to pool features
from (hence, pooling from a total of
$25\times 5 = 125$ frames per video).

\begin{table}[t]
\caption{Pooling from multiple crops of each frame
during testing of \methodTag{} model, compared to
single crop testing and baseline two-stream model
  (typically with 10-crop testing) averaged over 3 splits of HMDB51.
}\label{tab:multi-crop}
\tableSize{}
\centering
\begin{tabular}{lllll}
\toprule
  Method    & Spatial & Temporal & Combined \\
\midrule
  Two Stream~\cite{WangL_15_GoodPrac,WangX_16a}\footnote{Reported on \url{http://www.robots.ox.ac.uk/~vgg/software/two_stream_action/}} & 42.2 & 55.0 & 58.5 & \\
  \methodTag{} (Single-Crop) & {\bf 50.0} & 59.1 & 66.3 \\
  \methodTag{} (Multi-Crop) & {\bf 50.0} & {\bf 59.8} & {\bf 66.9} \\
\bottomrule
\end{tabular}
\end{table}

\section{Combination with iDT}
Combination with iDT for all datasets was done using late fusion, by weighted 
average of the iDT based score values for each class for each video, with the scores
predicted by our model. The iDT scores were obtained from the authors of~\cite{Varol_16}
and~\cite{charades} respectively.

\section{Comparison of baseline two-stream architecture with our approach}

\subsection{Confusion Matrix}
\begin{figure*}
    \centering
    \includegraphics[width=\linewidth]{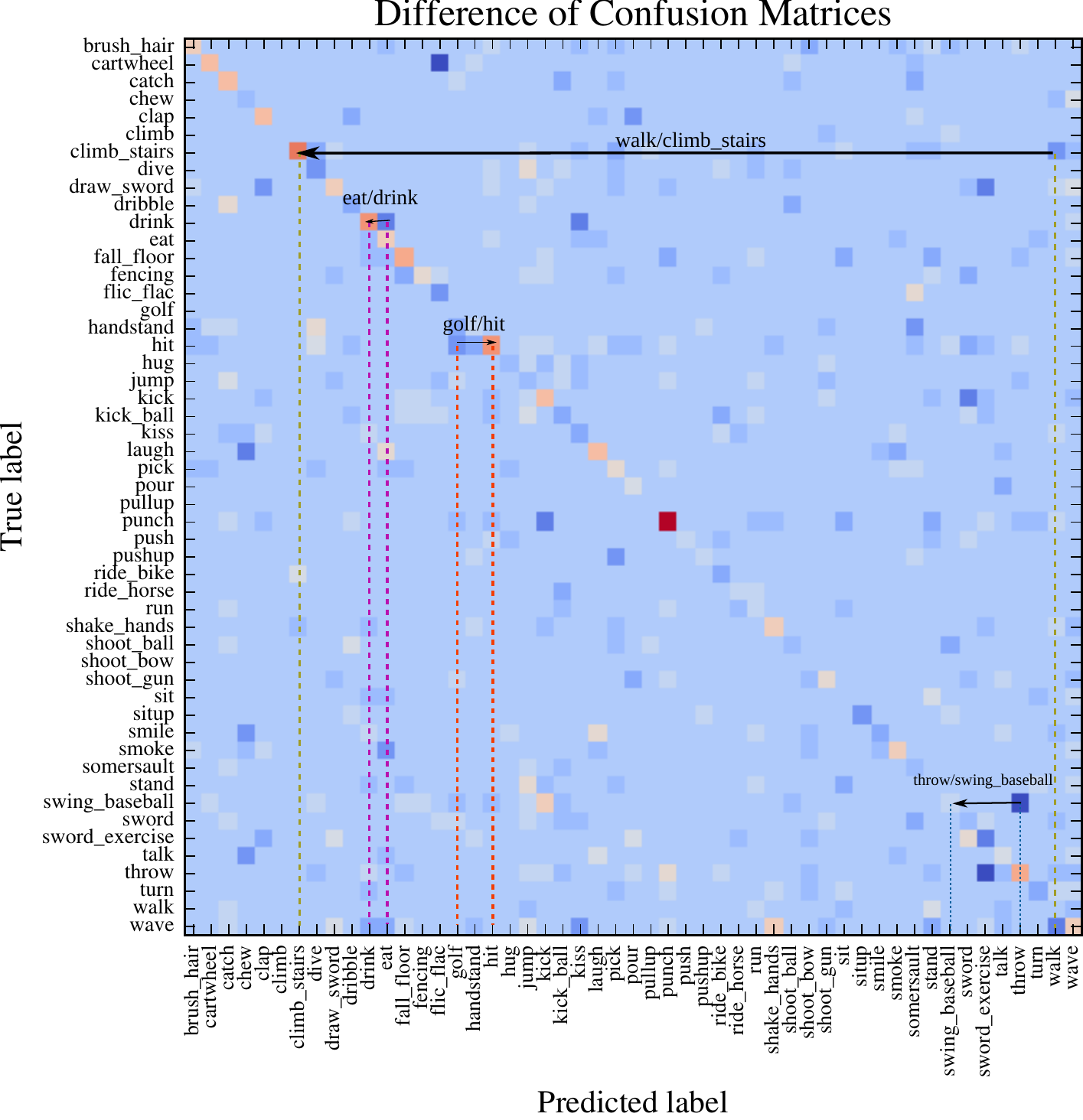}
    \caption{Difference of confusion matrices for classification on HMDB51 split 1 test set
    using our \methodTag{} representation versus the two-stream model.
    Red denotes positive changes and blue negative,
    so for improvements we would expect reds on the diagonal
    and blues off-diagonal.
    Compared to two-stream results, we see largest 
    improvement for classes such as `hit', which was earlier confused with `golf';
    or `climb\_stairs' which was confused with `walk'.}\label{fig:conf_mat_compare}
\end{figure*}

We visualize the classes for which we obtained the highest
improvements compared to the standard two-stream model.
We see the largest gains for classes such as `climb\_stairs',
which was earlier confused with `walk',
`hit', often confused with `golf'; and `drink', 
often confused with `eat' or `kiss'. This is expected,
as these classes are easy to confuse 
when averaging only global appearance and motion features.
Our model, on the other hand, can hone in on local appearance or motion features, such as a golf
club, making it easier to disambiguate actions such as `golf' from similar actions such
as `hit'.
The complete confusion matrix is in Figure~\ref{fig:conf_mat_compare}.

\subsection{Most Improved Videos}
Next, we visualize some videos that were incorrectly classified by the two-stream model, but 
are correctly classified by our model. We sort these videos by the classifier score with \methodTag{}, and show the top few in Table~\ref{fig:video_compare}.
The green label is our prediction, red is prediction by the two-stream model.

\begin{table*}[]
\centering
\caption{Some video predictions that were corrected by our method (red: prediction using two-stream,
green: ours).
For most of these cases, we see either that the action is hard to estimate from local evidence
  from a single frame, such as chew versus talk in (b); or that we need to focus on
specific visual features that can help disambiguate similar actions, such as existance
  of a glass for drinking in (a)}
\label{fig:video_compare}
\begin{tabular}{m{0.005\linewidth}m{0.8\linewidth}m{0.195\linewidth}}
  (a) & \includegraphics[width=\linewidth]{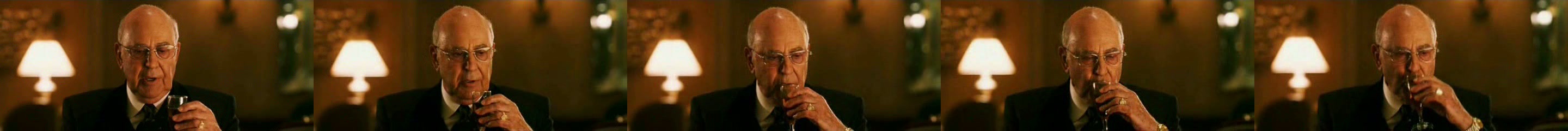} & \oldnew{eat}{drink} \\
  (b) & \includegraphics[width=\linewidth]{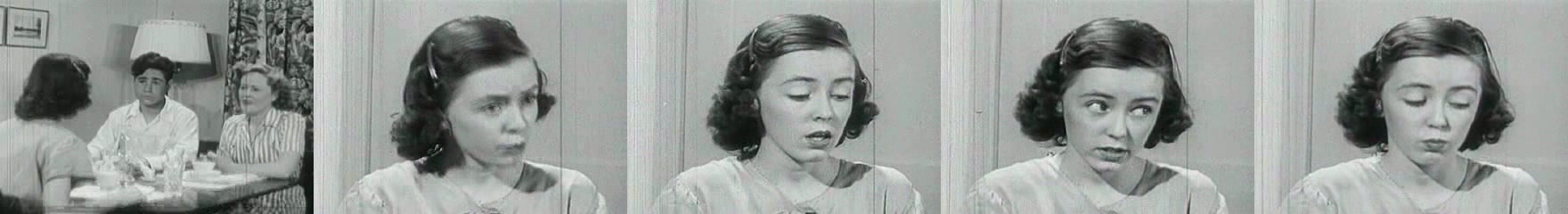} & \oldnew{chew}{talk} \\
  (c) & \includegraphics[width=\linewidth]{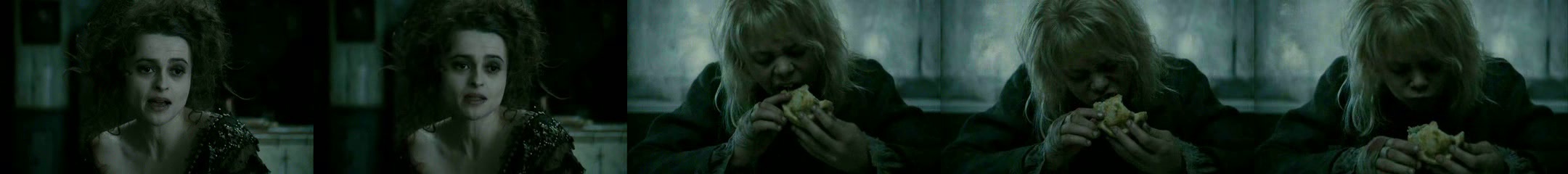} & \oldnew{drink}{eat} \\
	  (d) & \includegraphics[width=\linewidth]{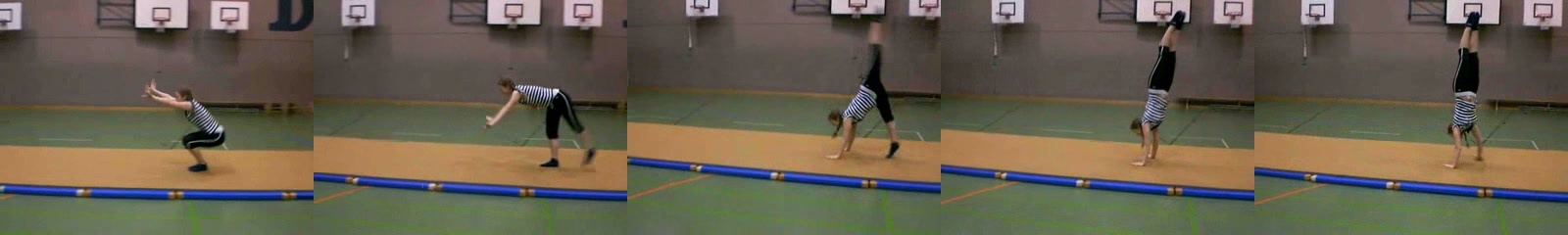} & \oldnew{somersault}{handstand} \\
  (e) & \includegraphics[width=\linewidth]{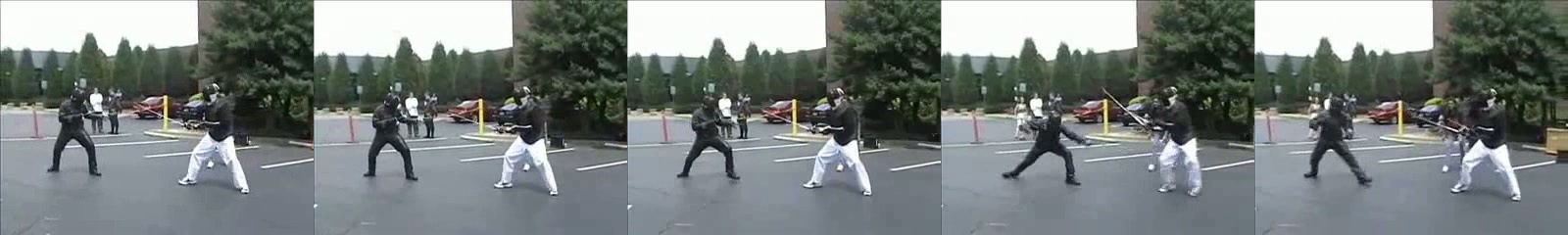} & \oldnew{sword}{fencing} \\
  (f) & \includegraphics[width=\linewidth]{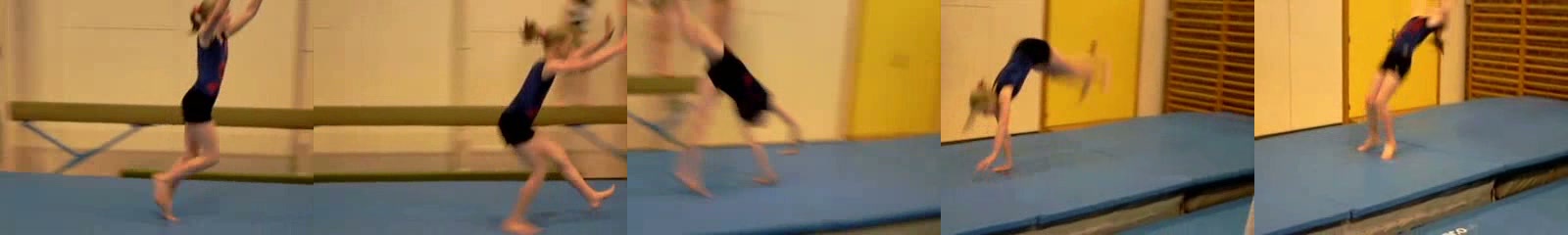} & \oldnew{flic\_flac}{cartwheel} \\
  (g) & \includegraphics[width=\linewidth]{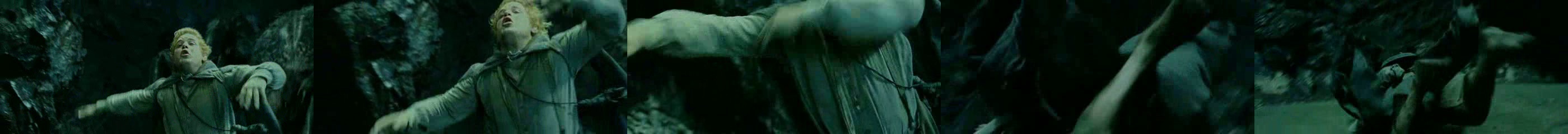} & \oldnew{punch}{fall\_floor} \\
  (h) & \includegraphics[width=\linewidth]{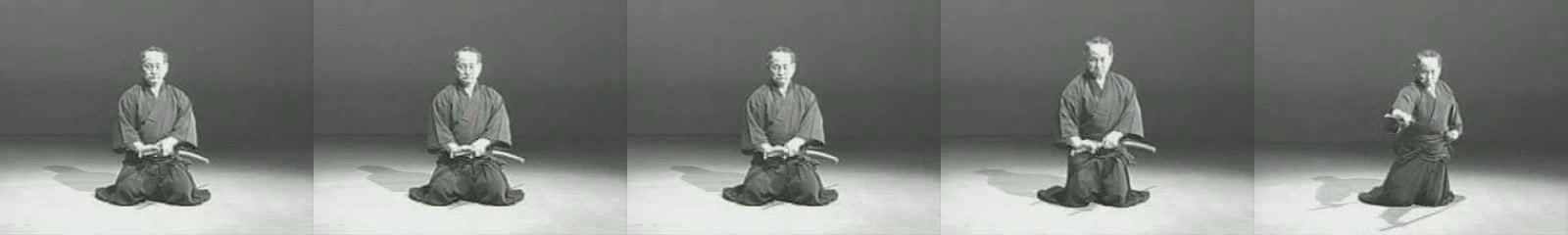} & \oldnew{sword\_excercise}{draw\_sword} \\
  (i) & \includegraphics[width=\linewidth]{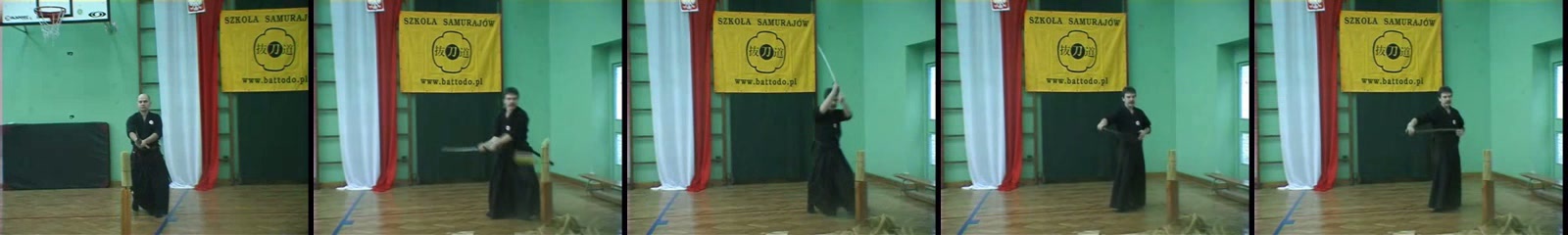} & \oldnew{draw\_sword}{sword\_excercise} \\
	  (j) & \includegraphics[width=\linewidth]{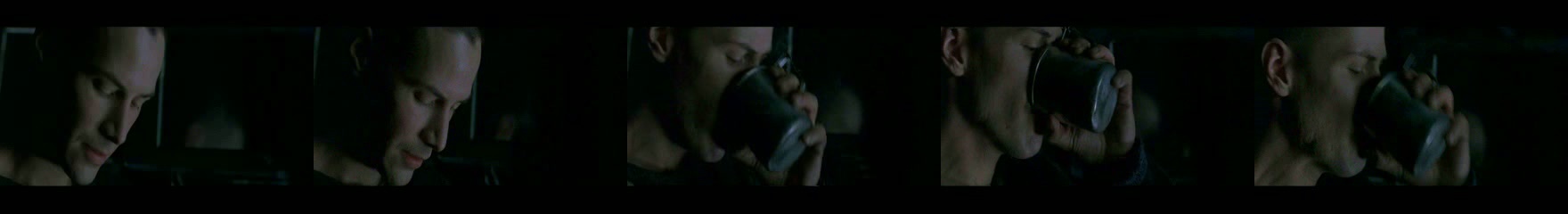} & \oldnew{kiss}{drink} \\
  (k) & \includegraphics[width=\linewidth]{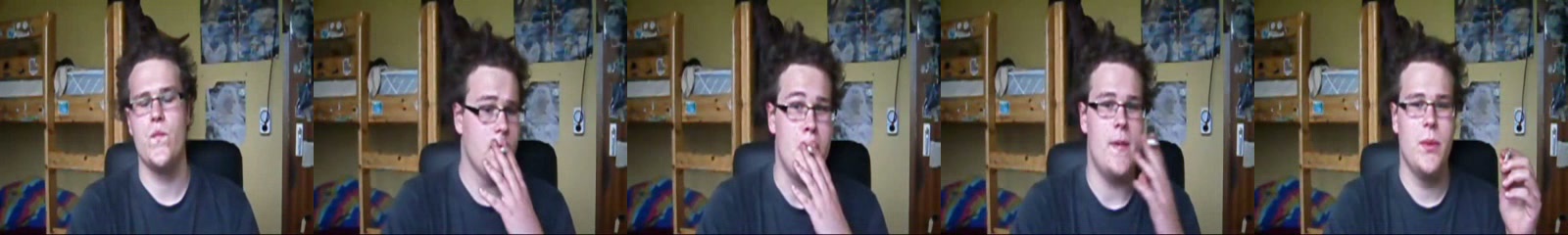} & \oldnew{eat}{smoke} \\
\end{tabular}
\end{table*}

\section{Visualization of action words}
In the attached video\footnote{\url{https://youtu.be/wVde6BPVUM0}}, we show a visualization of the various `action words' learned by our method, as well as our interpretation of those words, based on looking at randomly picked videos.
The highlighted regions in the videos correspond to the center of the receptive
field of conv5\_3 neurons, which are assigned to a specific cluster center (`action word').

\section{More Analysis}

\subsection{Combining Motion and Appearance}

\begin{figure}
    \centering
    \includegraphics[width=\linewidth]{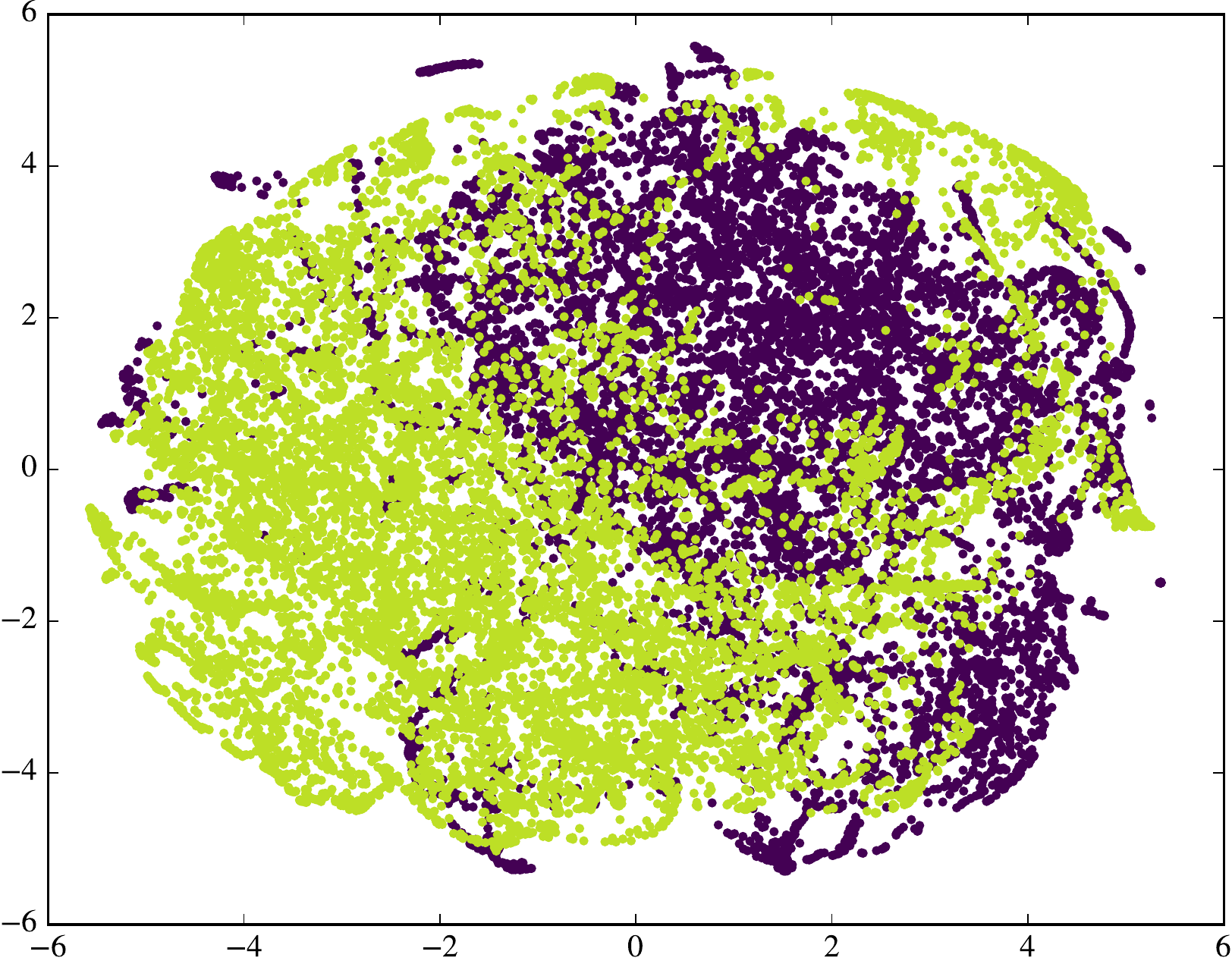}
    \caption{tSNE embedding for L2 normalized conv5 features of corresponding frames from appearance and motion streams, for 1000 randomly sampled frames.
    Though there is some mixing, the motion and appearance features occupy mostly distinct parts of the feature space, suggesting that a late fusion of ActionVLAD representations trained independently on both modalities will perform well. This is confirmed by our experiments
    in Table~\ref{tab:ablative-flow-rgb} in paper.}\label{fig:sup:conv5}
\end{figure}

Figure~\ref{fig:sup:conv5} shows a tSNE embedding plot for L2 normalized conv5 features from appearance and motion streams. The clear separation indicates there is potentially
complementary information and supports our results of late-fusing the two streams.

\section{Experimental details: Charades}

Since Charades~\cite{charades} is an untrimmed dataset as opposed to trimmed datasets like HMDB and UCF, we
follow a slightly different paradigm for training and testing. The training for Charades is done
over the trimmed action boundaries provided in the dataset, assuming each boundary as a 
separate trimmed clip. At test time, the video is processed as a whole, without considering
the boundary information. We extract 25 frames from the complete video, feed it
to two-stream or \methodTag{} models and extract the final layer features which are then
evaluated using the provided mAP/wAP evaluation scripts. However, note that we did not use 
flow information for Charades; both two-stream and \methodTag{} models were trained
only using RGB information, over a base BN-inception~\cite{WangL_16a} network.
 
\end{document}